\newcommand{\CLASSINPUTtoptextmargin}{1in}
\newcommand{\CLASSINPUTbottomtextmargin}{1.1in}

\documentclass[conference]{IEEEtran}
\IEEEoverridecommandlockouts
\usepackage{cite}
\usepackage{amsmath,amssymb,amsfonts}
\usepackage{algorithmic}
\usepackage{graphicx}
\usepackage{textcomp}
\usepackage{xcolor}
\def\BibTeX{{\rm B\kern-.05em{\sc i\kern-.025em b}\kern-.08em
    T\kern-.1667em\lower.7ex\hbox{E}\kern-.125emX}}
\begin{document}

\title{Auditing Facial Emotion Recognition Datasets for Posed Expressions and Racial Bias
}

\author{\IEEEauthorblockN{Rina Khan}
\IEEEauthorblockA{\textit{School of Computing} \\
\textit{Queen's University}\\
Kingston, Ontario, Canada \\
rina.khan@queensu.ca}
\and
\IEEEauthorblockN{Catherine Stinson}
\IEEEauthorblockA{\textit{School of Computing and} \\
\textit{Department of Philosophy} \\
\textit{Queen's University}\\
Kingston, Ontario, Canada \\
c.stinson@queensu.ca}
}

\maketitle

\begin{abstract}
Facial expression recognition (FER) algorithms classify facial expressions into emotions such as happy, sad, or angry. An evaluative challenge facing FER algorithms is the fall in performance when detecting spontaneous expressions compared to posed expressions. An ethical (and evaluative) challenge facing FER algorithms is that they tend to perform poorly for people of some races and skin colors. These challenges are linked to the data collection practices employed in the creation of FER datasets. In this study, we audit two state-of-the-art FER datasets. We take random samples from each dataset and examine whether images are spontaneous or posed. In doing so, we propose a methodology for identifying spontaneous or posed images. We discover a significant number of images that were posed in the datasets purporting to consist of in-the-wild images. Since performance of FER models vary between spontaneous and posed images, the performance of models trained on these datasets will not represent the true performance if such models were to be deployed in in-the-wild applications. We also observe the skin color of individuals in the samples, and test three models trained on each of the datasets to predict facial expressions of people from various races and skin tones. We find that the FER models audited were more likely to predict people labeled as not white or determined to have dark skin as showing a negative emotion such as anger or sadness even when they were smiling. This bias makes such models prone to perpetuate harm in real life applications.
\end{abstract}


\section{Introduction}
The way that facial expressions are used to express emotions in social interations has long been studied. People use facial expressions ubiquitously, such as smiling to greet others, and frowning to express dissatisfaction. Ancient philosophers and thinkers ranging from Aristotle to Zeng Guofan believed facial expressions to reveal the inner workings of the human mind. Today, automating the process of recognizing and interpreting facial expressions has become a key research area in computer vision. Computer scientists pursue facial expression recognition (FER) due to its potential applications to human computer interaction.

An important question regarding emotion recognition is whether our facial expressions truly indicate our inner emotional states. Paul Ekman \cite{ekman1971} proposed that there are six basic emotions universal across cultures, which include happiness, surprise, anger, fear, disgust and sadness. These emotions are widely thought to be revealed through facial expressions subconciously. Ekman's theory of emotion has been hugely influential and is widely accepted by FER researchers. Ekman's work is, however, not without criticism, with many scholars questioning whether facial expressions truly express innermost emotions or if they are dependant on social context. Quantative research such as Crivelli et al. \cite{crivelli2015} suggests that facial expressions are an unreliable indicator of inner emotion. Another theory that has become popular among newer FER researchers is Fridlund's Behavorial Ecology Theory \cite{fridlund1994humanfacialexpression}. Fridlund proposed that facial expressions are socially motivated and performative in nature; they represent what we want to express in a social context \cite{fridlund1994humanfacialexpression}.

This however raises some important questions about posed expressions. If expressions can be posed for social interactions, how do posed expressions differ from spontaneous ones? Some researchers have attempted to distinguish between spontaneous and posed expression, but this has proven to be a challenging task. A survey conducted by Jia et al. \cite{jiasvp} on various spontaneous vs posed distinction methods highlight the difficulty in distinguishing between posed and spontaneous expressions, particularly the gap in performance between detecting posed expressions and spontaneous ones.

This is the first audit we are aware of that seeks to verify the presence and proportion of posed versus genuine expressions in FER datasets. If there are a significant number of posed expressions in FER datasets, that could cause problems where FER applications are employed `in the wild'. Both general performance could be affected, and biases could be introduced given that performance of emotion expressions varies by culture. There is not even an established methodology for distinguishing between posed and spontaneous expressions \cite{jiasvp}. Here we fill that gap by proposing a methodology for detecting posed expressions. This new methodology draws on existing work, such as it is, and establishes new standards. We then employ this new methodology to conduct an audit on samples taken from two state-of-the-art FER image datasets, AffectNet \cite{affectnet} and RAF-DB \cite{RAFDB}.

There is prior reason to suspect that FER algorithms may exhibit biases against people of dark skin color and non-white races. Facial recognition algorithms have been shown to perform worse on dark skinned faces \cite{gendershades}. Lauren Rhue also performed an audit of commercial FER models using a small test dataset and binary race labels \cite{rhue2018}. No fairness audit on computer vision algorithms has analyzed differences in performance between more than two observed races. To address the lack of a comprehensive FER fairness audit, we audited two state-of-the-art models trained on AffectNet and RAF-DB respectively. We tested the models on FairFace \cite{fairface}, a large dataset with labels for observed race.

\section{Background}

\subsection{FER Datasets}

Affective computing is a domain of research that focuses on people's emotions while they interact with a computer system \cite{kolakowaska2014booksection}. Affective computing aims to be able to recognize emotions of users and influence them in order improve production satisfaction, productivity, and ease of accessing automated services. Recognition of emotion from faces is important to researchers working with real time image data. Many methods have been suggested  to model facial expressions. Ekman \cite{ekman1971} suggested a categorical system, where six basic emotions had visually distinct facial expressions associated to them. Ekman further proposed the facial action coding system, which assigned action units (AUs) to certain facial movements  \cite{ekmanfacs1978}. Russel proposed a multidimensional model involving continuous ranges of valence (how positive or negative a response is) and arousal (intensity of response) \cite{russel1980}. 

Datasets studying facial expressions were created as early as 1998, with the creation of JAFFE \cite{jaffe}, where ten Japanese women where asked to perform six different facial expressions. Other datasets created around this time such as the Cohn-Kanade dataset \cite{cohnkanade}, which also used controlled environments where participants were asked to perform expressions. Ekman's categorical model is widely used to provide emotion labels to facial expressions \cite{ekman1971}. Sometimes AUs are used in the labelling process, and then mapped to emotion categories. For example AU6 (raised cheeks) and AU12 (curled lips) indicate a happy face. The valence and arousal model is used much less due to the difficulty of labelling large sets of data.

Some limitation of using datasets developed in controlled settings are that they tend to include only posed expressions, existing ones are limited in the diversity of their data, and they tend to be small. In the 2010s, researchers found alternative methods for collecting more diverse, naturalistic data in larger quantities. The Acted Facial Expressions in the Wild dataset (AFEW) \cite{afew} collected videos from 54 movies. Later datasets also also use this method, such as Dynamic Facial Expressions in the Wild (DFEW) \cite{dfew}, which contains 16372 video clips from movies. FERV39K used videos from multiple sources such as film, TV, talk shows and interviews. Other researchers found a vast collection of images could be obtained through querying specific emotion-related keywords from the internet. The FER2013 dataset was introduced for the ICML 2013 Challenges in Representational Learning \cite{fer2013}. Similarly datasets would soon follow, such as EmotioNet \cite{emotionet} and AffWild \cite{affwild}. Some datasets use more creative methods of data collection, such as SASE-FE database \cite{kulkarni2018} from GoPro video recordings of 54 participants.

We selected two images databases that were collected through querying the internet, AffectNet \cite{affectnet} and RAFDB \cite{RAFDB}. We limited our analysis to image FER databases, rather than video for this analysis. AffectNet and RAFDB are two recent databases with a large number of annotated images. AffectNet contains about 300,000 images annotated manually with 8 emotion categories - Neutral, Happy, Sad, Surprise, Fear, Anger, Disgust, Contempt. RAFDB contains 29,672 images with 7 emotion categories - suprised, fearful, disgusted, happy, sad, angry and neutral.

\subsection{Spontaneous and Posed Expressions}

As affect theory developed, many researchers began exploring whether posed expressions were distinct from genuine ones. Early research in 1990s found that "smiles of enjoyment" were distinct from general smiles \cite{frank1993}. This was further investigated by Karen et al. \cite{karen2005} who discovered that posed and genuine smiles differed in facial movement and timing. Spontaneous expressions in these studies were identified through external stimuli, such as video clips and photos. Later, datasets were created with distinct spontaneous and posed emotions. Most datasets focused on smiles such as extended Cohn-Kanade (CK+) \cite{ckplus}, UvA-NEMO \cite{uvanemo} and  MAHNOB \cite{mahnob}. Smiles were a focus as it is the easiest for participants to pose leading to easier data collection. As a result, research on the distinction between genuine and posed exppressions were most extensively done on smiles. An example of this is Bogodistov and Dost \cite{bogodistov2017} who detect a genuine or Duchenne smile using the facial action coding system by the presence of AU6 (raised cheeks). Spontaneous expressions for other emotions such as surprise and pain have also been investigated \cite{bartlett2006, walter2013}, exclusively with video samples. Despite many researchers conducting studies in spontaneous and posed expression detection, there is no consistent methodology for detection of posed expressions, nor any comprehensive datasets created for this purpose \cite{jiasvp}.

Acted expressions in movies are often included in FER research in place of truly spontaneous expressions. For instance, Dhall et al. argue for using movie scenes in AFEW in lieu of real life data as an acceptable compromise, due to a lack of data. They claim that performances by skilled actors are an acceptable proxy for genuinely spontaneous emotional expressions \cite{afew}. Other research suggests non-actors are better suited for emotion studies than actors in terms of how genuine they are perceived to be \cite{jurgens2015}. There are other complications involved in using acted expressions in movies as a proxy for spontaneous expressions, such as the performance of gender stereotypes \cite{wallbott} and racial stereotypes \cite{mary2011} in film.

\subsection{Algorithmic Audits of Computer Vision Algorithms and Fairness Datasets}

With the rise in commercial and non-commercial uses of AI, algorithmic fairness has become a major area of concern \cite{algorithmsofoppression, weaponsofmathdestruction}. Studies have performed benchmarking and other analyses to investigate fairness in algorithmic systems. These include racial and gender bias in sentiment analysis \cite{kiritchenko2018} and generative models \cite{mcduff2019}. Computer vision has received particular interest, with image based gender classification shown to be biased against the intersection of darker and female faces \cite{gendershades}. Image classification has also been found to be less likely to predict STEM-related keywords for feminine faces \cite{genderslopes}.

A common limitation of many of these studies is that racial classifications are limited to just two or a few races, due to the lack of datasets with balanced racial labels at the time the studies were conducted. Popular image databases such as ImageNet \cite{imagenet} have significant disparities in representation and it is difficult to remedy due to the difficulty of collecting huge amounts of data from the internet while maintaining diversity \cite{shankar2017}. Auditors sometimes overcome this by collecting their own datasets. The gender classification audit collected a dataset of parliamentary figures from three African and three European countries \cite{gendershades}. Joo and Kärkkäinen used synthesized images to create a diverse collection of images \cite{genderslopes}.

Researchers began creating fairness datasets to fill the gap of a lack of representation in large datasets. One such dataset is UTKFace which includes 20,000 images with age, gender and racial labels. Race is annotated by observers as White, Asian, Black, Indian and Other \cite{utkface}. This dataset still had a significantly higher number of images labeled as White than other labels and "Other" is an extremely broad label that serves little practical use. Fairface \cite{fairface} is another such dataset that has more balanced race labels. Fairface contains 108,000 images with observed races that include White, Middle-Eastern, East Asian, Southeast Asian, Black, Indian and Latino.

\section{Methodology}

\subsection{Dataset Sampling}

We selected our two FER datasets to audit - AffectNet \cite{affectnet} and RAFDB \cite{RAFDB}. Our justification for selecting the two was that we were interested particularly in image datasets that had a large number of images and are in-the-wild. AffectNet and RAF-DB are two recent image based FER datasets that have gained a large amount of popularity and have been stated as state-of-the-art \cite{relativeuncertainitylearning2021}. The full AffectNet dataset has about a million images, where a subset of 291,651 images are manually labeled. We were more interested in examining this subset as it is more representative of how the authors and annotators intended the dataset to be. We randomly sampled 381 images from the dataset. The sample size was calculated to be representative of the dataset with 95\% confidence level and 5\% margin of error. RAF-DB contains 29,672 total annotated images, but these are separated into two sets, basic and compound, with images repeating between sets. The compound set includes the combination of the top 2 labels from annotators to provide a compound emotion. We chose to use the basic subset, which has the top label, as it is consistent with AffectNet which allows for a simpler auditing method. The basic set of RAF-DB contains 15,339 images. Using the same sampling parameters for calculating a representative sample size, we randomly sampled 380 images. Both datasets had a large imbalance in labels with far more images labeled as ``Happy". To ensure a more diverse sample is collected, we sample equal numbers of each emotion label. We then further analyze the samples, observing the whether they were spontaneous or posed, skin tone of the people pictured, and facial expressions.

\subsection{Spontaneous versus Posed}\label{sec:svp-detection}

We found it challenging to adopt an established methodology for identifying spontaneous versus genuine expressions,  Existing studies  do not cover the full range of emotions present in the datasets and there is no consistent method of identifying posed or spontaneous emotions. Smiles are very well studied however, and we had confidence in adopting methods of identifying posed smiles. We chose to adopt Bogodistov and Dost's FACS based method \cite{bogodistov2017}, which identified a genuine smile as consisting of AU6 (raised cheek), AU12 (curled lip) and AU25 (parted lips). A posed smile is identified as AU12 and AU25 but with the absence of AU6.

When the person in the image is not smiling, in the absence of a consistent technique of judging the expression, we considered some alternate factors that would give away a posed image. We considered a recognizable actor in a movie or TV scene to be performing a posed expression. While actors attempt to mimic genuine expressions, performance of an emotion is not the same as experiencing the said emotion. Furthermore, acting involves a degree of playing up to the camera and audience, where actors have to exaggerate their performances to clearly communicate their intended emotion to the audience. Acted expressions are at the least an unreliable source for representing genuine expressions.

We also consider the background of the image. It is common for photographers to pose models on plain mono-color backgrounds to use their images as stock images. Genuine expressions in the wild are much more likely to have an environment in the background. While of course there may be images of people posed with an environment in the background, by identifying cases of plain mono-color backgrounds we can identify images where we have a high confidence of the image being posed.

Finally we identify if a person is looking straight at the camera and if they are very well lit without being out in the sunlight. Someone looking straight at the camera has some likelihood of posing for a photographer. A very well lit image that is not natural sunlight may be artificial lighting set up by a photographer. These factors in isolation do not give a high confidence of an image being, someone could be showing a genuine expression while coincidentally staring at a camera. However, if a person is staring straight while in a very well-lit environment, we can get a high confidence of the image being posed.

\subsection{Skin Tone}

Because of the complexity of race and impossibility of determining it reliably from images, we used skin tone as a proxy (however imperfect) for race. We discuss this further in Section \ref{sec:race}. Skin tone labels were assigned according to Fitzpatrick six point labelling system, which is used by dermatologists to determine risk of skin cancer \cite{Fitzpatrick1988}. We separated the skin tones labels into three groups - Group 1 representing lightest skin tones (skin tones I and II), Group 2 (skin tones III and IV) and Group 3 representing darkest skin tones (skin tones V and VI).

\subsection{Fairness Dataset}

We used FairFace \cite{fairface} as our test dataset. FairFace is by far the largest fairness dataset with the most number of observed race labels. The labels are also well balanced in comparison to other similar datasets.

\subsection{Model Selection and Inference}
 To assess which models to test, we searched for the best performing models on the website PapersWithCode \cite{papers_with_code_papers_2022}. Models where extra training data is used are excluded since we want to observe any relationships between models and the datasets they are trained on. The models selected are summarized in Table \ref{tab:models-selected}.

\begin{table}
  \caption{Models selected for testing}
  \label{tab:models-selected}
  \begin{tabular}{ccl}
    \hline
    Dataset&Model&Accuracy\\
    \hline
    RAF-DB& Relative Uncertainity Learning \cite{relativeuncertainitylearning2021} & 88.98\% \\
    AffectNet & Multi-task EfficientNet-B2 \cite{multitaskefficientnet2022} & 66.29\% \\
  \hline
\end{tabular}
\end{table}

The models are then evaluated for bias using datasets annotated with skin tone, race or ethnicity. We selected a large dataset with observed race labels, and used the FER models on the images in the dataset to predict their emotion. Once inference is run, we further sample from the inferences to observe if trends between predicted labels and skin tone exists. We aim to balance our samples by observed race and predicted emotion, selecting 50 samples for each permutation. If there are less than 50 samples for a specific permutation, we select as many as there are.

\subsection{Smiles and Neutral expressions}

We labelled if a person in an image was observed to be smiling or having a neutral expression for both samples collected from FER datasets and predictions on FairFace. We only labelled these two expressions because these were the easiest to identify. Smiles are identified by the presence of AU12 and AU25. Neutral expressions are not defined by the FACS. We instead employ a definition used by Adams et al., which is when a face is ``completely devoid of overt muscle movement leading to expression" \cite{adams_emotion_2012}. In terms of FACS, we interpret this as a lack of any identifiable AUs.

\subsection{Some considerations on race and skin tone} \label{sec:race}
Race is a useful parameter for algorithmic auditing, particularly for bias in surveillance applications. However defining race as a variable is complicated, because race is multidimensional and socially constructed, with no stable definition. We turn to Hanna et al. \cite{HannaRace2020}, who propose that the appropriate dimension of race is based on how race data is collected. In FairFace, race labels are obtained from human annotators observing images. Dimensions like self classification (the race we check on official forms), racial identity (subjective self identification), reflected race (race we believe others observe us as) and phenotype (appearance based markers of race such as skin color) \cite{HannaRace2020, RothMultipleDimensions} are not reflected in FairFace. ``Observed race" will be used here-on-after to describe as that is a term that more accurately describes the labels obtained through observation.

Because judging race from someone's image is a very difficult task, Gebru and Buolamwini used skin tone as a variable for their audits \cite{gendershades}. There are limitations to observing skin tone in naturalistic settings, where lighting varies significantly from image to image. Despite this, observed skin tone can be useful to assess what the FER model ``sees", and thus we use it here too.

\section{Results}

\subsection{Spontaneous versus Posed}

\begin{figure}[h]
  \centering
  \includegraphics[width=\linewidth]{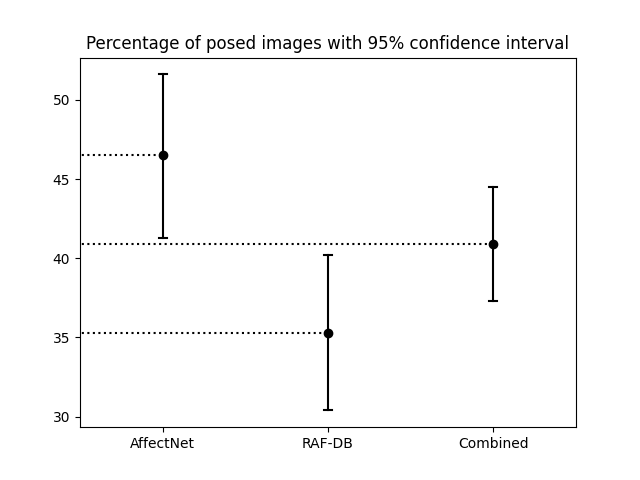}
  \caption{Percentage of posed images in datasets with 95\% confidence interval. AffectNet: 46.5\% (CI: 41.3\%, 51.6\%), RAF-DB: 35.3\% (CI: 30.4\%, 40.21\%), Combined: 40.9\% (CI: 41.3\%, 51.6\%)}
  \label{fig:svp-confidence-intervals}
\end{figure}

When analyzing samples from datasets, we found a large number of images to be posed. 46.5\% of images in AffectNet, 35.3\% in RAF-DB and 40.9\% of the combined set were posed. Figure \ref{fig:svp-confidence-intervals} shows these values with upper and lower bounds at 95\% confidence interval. We can establish a null hypothesis of observing zero posed images in a completely wild setting with $\alpha = 0.01$. The results are statistically significant, Pearson chi-square: $\chi^2 (1; N = 761) = 385.2, p < 0.01$.

\subsection{Model inference}

We predicted emotions of images in Fairface using two models, Relative Uncertainity Learning (RUL) trained on RAF-DB and Multi-task EffecientNet B2 (MENet) trained on AffectNet. We sampled the predictions leading to a distribution of samples by race in table \ref{tab:sample-model-preds}.

\begin{table}
    \centering
    \begin{tabular}{cccc}
        \hline
        Observed Race & RUL & MENet & Total \\
        \hline
        East Asian & 162 & 200 & 362 \\
        Indian & 175 & 200 & 375 \\
        Black & 161 & 200 & 361 \\
        White & 201 & 200 & 401 \\
        Middle Eastern & 193 & 202 & 395 \\
        Latino/Hispanic & 192 & 201 & 393 \\
        Southest Asian & 163 & 200 & 363 \\
        \hline
        Total & 1247 & 1403 & 2650 \\
        \hline
    \end{tabular}
    \caption{Number of samples collected from model predictions on Fairface}
    \label{tab:sample-model-preds}
\end{table}

The overall predictions skew towards `Happiness' for both models but this also accurately reflects that Fairface has mostly images of people smiling. Observing samples for intersection of emotion prediction and observed race reveal some interesting findings. We observed that a significant portion of negative emotion predictions were smiling. We considered anger, sadness, contempt and disgust to be negative emotions. Among the various samples, people of observed race other than white had a higher proportion of smiling faces within the negative prediction set. A similar trend was observed with neutral faces.

\begin{figure}
    \centering
    \includegraphics[width=\linewidth]{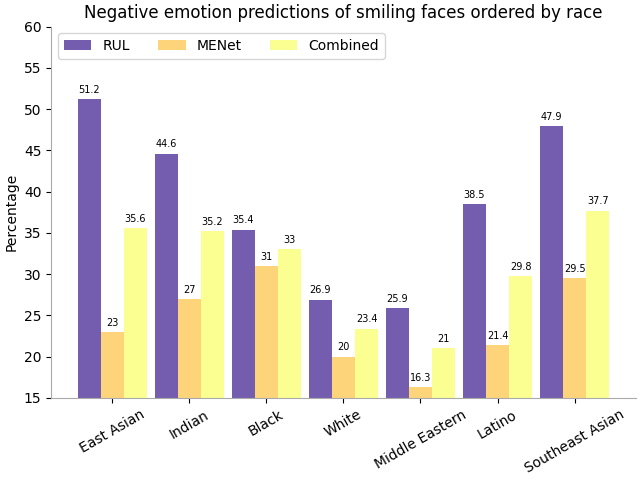}
    \caption{Percentage of negative emotion predictions in samples that are observed as smiling sorted by observed race and FER model}
    \label{fig:negative-race-smile}
\end{figure}

Across both models, 23.4\% samples with negative predictions observed as White are smiling as opposed to 33\% who are seen as Black. This number is 35.6\% for East Asian, 35.2\% for Indian, 33\% for Black, 29.8\% for Latino and 37.7\% for Southeast Asian. This is visualized in Figure \ref{fig:negative-race-smile}. Assuming the model is unbiased, the number of smiling faces in the negative prediction set should be the same as for those observed as white. Using that as our null hypothesis, we calculate statistical significance using Pearson chi-square for the difference between each osbserved race and white in Table \ref{tab:chi-squared-smile}.

\begin{table}[]
    \centering
    \begin{tabular}{ccc}
    \hline
              & $\chi^2$ & Significant? \\
        \hline
        Black & $8.09, p < 0.05$ & Yes \\
        East Asian & $13.1, p < 0.05$ & Yes \\
        Southeast Asian & $17.8, p < 0.05$  & Yes \\
        Middle Eastern & $0.54, p > 0.05$ & No \\
        Latino & $3.7, p > 0.05$ & No \\
        Indian & $12.4, p < 0.05$ & Yes \\
        \hline
    \end{tabular}
    \caption{Pearson Chi-square statistic for smiling faces in negative prediction set by observed race in relation to null hypothesis that error rate is to be the same as White. $ \alpha = 0.05 $}\label{tab:chi-squared-smile}
\end{table}

\begin{table}[]
    \centering
    \begin{tabular}{ccc}
    \hline
              & $\chi^2$ & Significant? \\
        \hline
        Black & $26.2, p < 0.05$ & Yes \\
        East Asian & $17.5, p < 0.05$ & Yes \\
        Southeast Asian & $6.14, p < 0.05$  & Yes \\
        Middle Eastern & $27.8, p < 0.05$ & Yes \\
        Latino & $8.16, p < 0.05$ & Yes \\
        Indian & $34.6, p < 0.05$ & Yes \\       
    \hline
    \end{tabular}
    \caption{Pearson Chi-square statistic for neutral faces in negative prediction set by observed race in relation to null hypothesis that error rate is to be the same as White. $ \alpha = 0.05 $}
    \label{tab:chi-squared-neutral}
\end{table}

A similar trend was observed with neutral faces. 32.7\% of negative predictions seen as White had neutral faces, and this number was 51.2\% for Black, 51.4\% for Middle Eastern, 47.8\% for East Asian, 53.9\% for Indian, 42.7\% for Latino and 41.6\% for Southeast Asian. We again use a null hypothesis that if the model is unbiased, the number of neutral faces in the negative predictions would be the same as those observed as White. We calculate statistical significance for difference between each observed race and White in Table \ref{tab:chi-squared-neutral}.

\begin{figure}
    \centering
    \includegraphics[width=\linewidth]{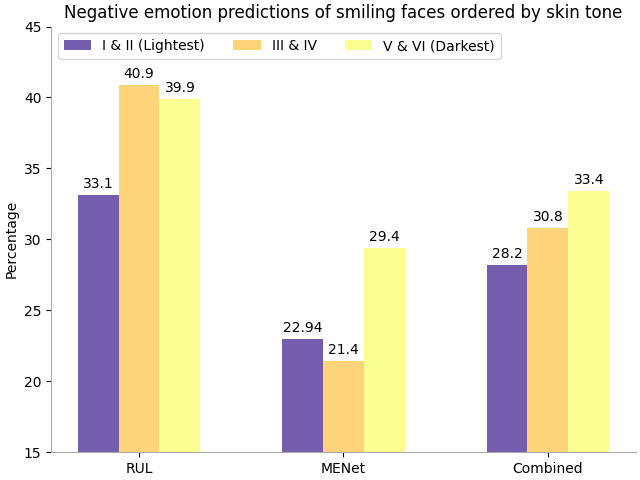}
    \caption{Percentage of negative emotion predictions in samples that are observed as having smiling faces sorted by skin tone and FER model}
    \label{fig:negative-skin-tone-smiling}
\end{figure}

Observing trends across skin tone also reveals a greater discrepency for darker skin. We visualized this by grouping skin tone labels into three categories, Skin tone I and II representing the lightest skin tone, skin tone III and IV representing darker skin tone and skin tone V and VI representing the darkest skin tone. This is shown in Figure \ref{fig:negative-skin-tone-smiling}.

\section{Discussion}

We observed bias against both dark skinned individuals and those seen as non-white in the form of higher error rates. Previous research hints that a large contributing factor to bias in AI is under-representation of certain groups in datasets. In this set of experiments, we did observe a correlation between the bias with under-representation in the dataset samples. Only 4\% of the sampled images were of the darkest skin tone. We were interested in checking for if the dataset annotations also demonstrated stereotypes or biases against the darker skin group. However, with only so few dark skinned people being present it was impossible to observed any statistically significant trends.

Racial bias being present in FER models is particularly unsettling given social context of emotion perception. Human emotion perception is thought to be instant and a natural process, however is also flawed \cite{Todorov2017}. Despite these flaws, legal actors such as judges nevertheless provide harsher sentences to defendants whose natural facial expression is seen as angry \cite{JPWilson2015, JPWilson2016}. Societal and racial biases also influence such judgements. Halberstad et al. have noted that in schools, Black children are more likely to be seen as angry than white children \cite{Halberstadt2022}.

There are some serious harms that could be inflicted should racial biases in FER models and datasets not be accounted for and resolved. Applications of FER technology involve automated interviews and crowd security. Many FER scholars primarily cite Ekman's theory of facial expression revealing our innermost emotions. This context can easily allow biased applications of FER models to be interpreted in a very harmful manner. For instance, a security application could observe a dark skinned person as more prone to anger and inherently more likely to commit a crime. We strongly encourage FER researchers going forward to re-evaluate this interpretation and instead adopt Fridlund's Behavior Ecology Theory. This would imply that FER technology would be a tool for social communication, used for reading intentionally presented facial expressions in a particular social context. This would also encourage adoption of FER technology for communication applications where we believe it is much better suited than security.

Our practices of collecting huge amounts of data for training machine learning models presents very difficult challenges when it comes to annotations. In the case of labeling facial expressions, it is almost impossible thousands of images in a manner that does not allow social and cultural biases to creep in. People have been shown to be more negative in judging facial expressions when they see the person as a racial ``other" \cite{EmotionalPerceptionRacialBias}. Cultural norms of facial expressions and emotion performance for one racial group vary from others \cite{barrett_emotional_2019}. Ideally, a team of diverse annotators could all take part in annotations. However, this is very difficult to achieve with the sheer volumes of annotations in any machine learning dataset.

The significant number of posed images in datasets pose a challenging problem as any environment an FER model would be deployed in would have a significant difference from its training data. In future work we would like to further investigate the effects of posed expressions on model performance. We would also like to analyze if posed expressions and racial bias is observed in video-based FER datasets and models.

\section{Conclusion}

The aim of this study was to provide a comprehensive audit of image FER datasets and models focused on spontaneous and posed emotion and racial bias. We looked at samples from two state-of-the-art and influential FER datasets and the best performing models trained on the datasets. We found a statistically significant number of posed images. Given that performance of models are different between spontaneous and posed expression, this presents a serious performance concern regarding the deployment of FER models in real world settings. We also observed that a larger number of people with darker skin tone and those seen as non-white were predicted to have negative emotions even when they were smiling or had a neutral expression. FER authors should be more cautious regarding the framing of FER technology, and instead present FER as a means to detect and transmit intentionally expressed social cues.

\bibliography{references}
\bibliographystyle{acm}

\end{document}